\documentclass[letterpaper]{article} 
\usepackage{aaai2027}  
\usepackage[hyphens]{url}  
\usepackage{graphicx} 
\usepackage{natbib}  
\usepackage{caption} 
\usepackage{algorithm}
\usepackage{algorithmic}
\usepackage{amsmath}
\usepackage{subcaption}
\usepackage{multirow}
\usepackage{xspace}
\usepackage{comment}
\usepackage{tabularray}
\usepackage{amssymb}
\usepackage{stmaryrd}
\usepackage{mathtools}
\usepackage{amsthm}
\usepackage[table]{xcolor}
\usepackage{tikz-cd}

\newcommand{\conv}[1]{$\llbracket #1 \rrbracket$}
\newcommand{\Nlean}{$\mathbb{N}$}
\newcommand{\defined}[1]{\textbf{#1}}
\theoremstyle{plain}
\newtheorem{theorem}{Theorem}

\newtheorem{lemma}{Lemma}
\newtheorem{corollary}{Corollary}
\theoremstyle{definition}
\newtheorem{definition}{Definition}

\theoremstyle{remark}

\theoremstyle{remark}

\DeclareCaptionType{listing}[Listing][List of Listings]

\usepackage{newfloat}
\usepackage{listings}
\DeclareCaptionStyle{ruled}{labelfont=normalfont,labelsep=colon,strut=off} 
\floatstyle{ruled}
\newfloat{listing}{tb}{lst}{}
\floatname{listing}{Listing}

\lstdefinelanguage{Lean}{
  morekeywords=[1]{theorem, lemma, def, example, induction, structure, abbrev, inductive},
  morekeywords=[2]{have, let, show, from, fun, match, with, cases, intro, exact, apply,
                by, obtain, rfl, sorry, where, if, then, Prop, false, true, false}
  sensitive=true,
  morecomment=[l]{--},
  morestring=[b]",
}

\lstdefinestyle{leanstyle}{
  language=Lean,
  basicstyle=\footnotesize\ttfamily,
  keywordstyle=[1]\color{blue!70!black}\bfseries,
  keywordstyle=[2]\color{green!70!black}\bfseries,
  commentstyle=\color{gray},
  stringstyle=\color{orange!80!black},
  showstringspaces=false,
  breaklines=true,
  frame=single,
  numbers=none,
  captionpos=b,     
  tabsize=2,
  mathescape=true,
  aboveskip=10pt,belowskip=5pt
}

\lstdefinestyle{leanstyle2}{
  language=Lean,
  basicstyle=\footnotesize\ttfamily,
  keywordstyle=[1]\color{blue!70!black}\bfseries,
  keywordstyle=[2]\color{green!70!black}\bfseries,
  commentstyle=\color{gray},
  stringstyle=\color{orange!80!black},
  showstringspaces=false,
  breaklines=true,
  frame=single,
  numbers=left,
  captionpos=b,     
  tabsize=2,
  mathescape=true,
  aboveskip=10pt,belowskip=5pt
}

\lstdefinelanguage{PDDL}{
  morekeywords=[1]{define, :domain, :objects, :init, :goal, problem},
  morekeywords=[2]{and},
  alsoletter={:},
  sensitive=true,
  morecomment=[l]{--},
  morestring=[b]",
}

\lstdefinestyle{pddlstyle}{
  language=PDDL,
  basicstyle=\footnotesize\ttfamily,
  keywordstyle=[1]\color{blue!70!black}\bfseries,
  keywordstyle=[2]\color{green!70!black}\bfseries,
  commentstyle=\color{gray},
  stringstyle=\color{orange!80!black},
  showstringspaces=false,
  breaklines=true,
  frame=single,
  numbers=none,
  captionpos=b,     
  tabsize=2,
  mathescape=true,
  aboveskip=10pt,belowskip=5pt
}
\usepackage{booktabs}

\title{Provably Complete Generalized Planning with LLMs}
\author{
    Katharina Stein\textsuperscript{\rm 1},
    Chaahat Jain\textsuperscript{\rm 1},
    J\"org Hoffmann\textsuperscript{\rm 1,2},
    Alexander Koller\textsuperscript{\rm 1}
}
\affiliations{
    \textsuperscript{\rm 1}Saarland Informatics Campus, Saarland University, Saarbr\"ucken, Germany\\
    \textsuperscript{\rm 2}German Research Center for Artificial Intelligence (DFKI), Saarbr\"ucken, Germany\\
    Correspondence: kstein@lst.uni-saarland.de

}

\begin{document}

\maketitle

\begin{abstract}
Generalized planning aims to compute a plan that solves all instances of a planning domain. Recent work has used LLMs to automatically generate and debug such generalized plans in the form of Python programs and achieved perfect test data coverage for several domains. However, whether these generalized plans are actually complete, i.e. solve all instances of the domain, could only be determined by manual evaluation. 
Here, we present an approach for automatically generating generalized plans in Lean together with proofs of their completeness relative to a specification of the domain constraints provided as input. We introduce a semantic-preserving PDDL-to-Lean conversion, and use an LLM to generate both the generalized plan and the formal proof that it solves every instance satisfying the domain constraints. The correctness of the completeness proof is determined by Lean's kernel. 
We evaluate our approach on 13 commonly used benchmark domains, using GPT-5.6-Sol as the LLM. For 12 of the domains we obtain generalized plans together with valid completeness proofs. 
%
This is a major advancement of the state of the art in automatic generalized-plan completeness proofs.
\end{abstract}


\section{Introduction}

\begin{figure}
    \centering
    \includegraphics[width=0.95\linewidth]{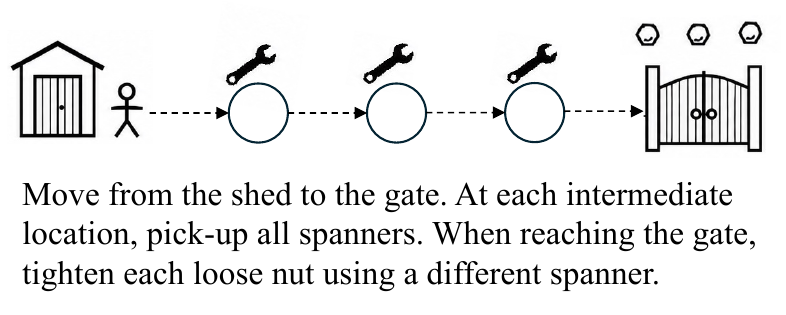}
    \caption{Illustration of an example planning task from the Spanner domain and a strategy for solving all Spanner tasks.}
    \label{fig:spanner_strat}
\end{figure}

Generalized planning aims to compute a plan that solve not only a single planning task, but all tasks from a given PDDL domain. Such a strategy will typically contain loops and conditionals to deal with different cases and sizes of tasks.
For example, in the Spanner domain from the IPC Learning track 2023 a man can walk between linked locations, pick up spanners and use a spanner to tighten a loose nut which makes the spanner unusable \cite{taitler-et-al-aimag2024}. As per the IPC problem generator, the links between locations form a one-way path from the initial location (a shed) to the location of all loose nuts (a gate), and all usable spanners are located between the shed and the gate. The goal is to tighten all loose nuts. Figure \ref{fig:spanner_strat} shows an example planning task and a strategy that generalizes to all tasks of the domain. 

\begin{figure}[t]
    \centering
    \includegraphics[width=0.75\linewidth]{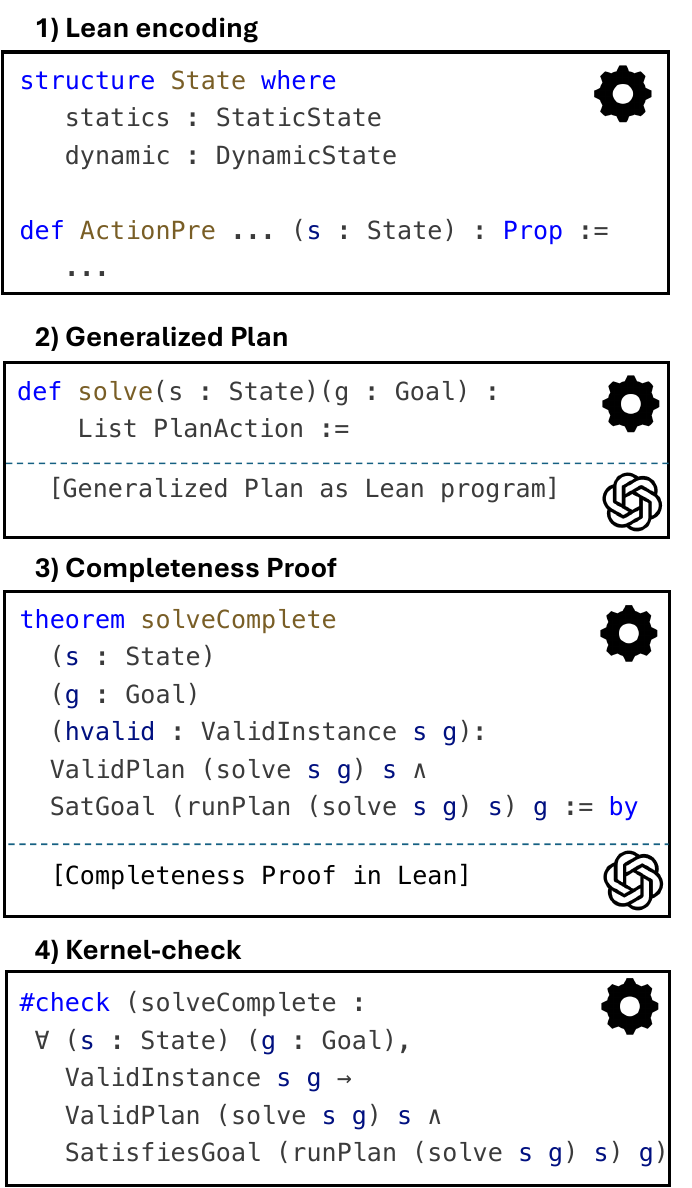}
    \caption{Illustrations of the main steps of our approach for obtaining generalized plans in Lean and proving their completeness.}
    \label{fig:approach}
\end{figure}

A variety of approaches for formalizing and obtaining generalized plans have been proposed, 
including representing generalized plans as synthesized programs with loops and conditionals \cite[e.g.][]{DBLP:journals/ai/SrivastavaIZ11, Jimenez_Segovia-Aguas_Jonsson_2019} or general policies \cite[e.g.][]{bonet-et-al-aaai2019, bonet-geffner-ijcai2018}.
Recently, there has been work on using Large Language Models (LLMs) to generate, and debug, generalized plans in the form of Python programs \cite{Silver_2024, stein2026improved}, with \citet{stein2026improved} achieving perfect (100$\%$) coverage on the test datasets for 14 benchmark domains.
Yet such empirical evidence is limited to a set of example tasks (the test dataset). Do the generalized plans actually solve \emph{all} tasks in the domains?

\citet{stein2026improved} showed through a manual analysis that this is indeed the case (similar manual analyses were done by \citet{drexler-et-al-icaps2022} and \citet{frances-et-al-aaai2021}). 
But how to lead this kind of proof automatically? \emph{Given the PDDL description of a domain, as well as some specification of the set of domain instances (tasks in the domain) that a generalized plan should solve, how to automatically obtain a generalized plan that provably solves all these instances?} We refer to this challenge as the problem of computing \emph{complete} generalized plans.

Earlier work on this challenge identified a restricted class of (so-called one-dimensional) planning problems where complete generalized plans can be found automatically using logical reasoning techniques \cite{hu-levesque-kr2010}.
Another strand of work has addressed the complementary challenge of, given a generalized plan, or as part of the computation of the generalized plan, determining a set of instances which the plan is guaranteed to solve \cite[e.g.][]{DBLP:journals/ai/SrivastavaIZ11,bonet-et-al-ijcai2019}. The key concept in the most recent work on this are so-called sound abstractions, which underlie the generalized plan and guarantee faithfulness to a concrete set of instances.
Recent work has shown how to automatically prove such abstractions sound, and therewith solve the challenge of proving a given generalized plan complete \cite{cui2023automatic}. The empirical success with this method is, however, limited to simple domains and/or instance sets, such as clearing a single block in Blocksworld, the IPC Gripper domain, and Logistics with a single vehicle.

In the present work, we leverage the power of LLMs to not only generate the generalized plans, but also the completeness proofs in a suitable formalization. 
Specifically, we prove completeness of LLM-generated generalized plans using Lean \cite{Lean}, a functional programming language and theorem prover. We generate the generalized plans directly in Lean, and automatically prove their completeness relative to \emph{domain constraints} formalizing the set of instances.

We automatically generate Lean code from the PDDL enforcing the domain semantics. Following \citet{Silver_2024} and \citet{stein2026improved}, we let the LLM generate the generalized plan as a program that takes a problem instance as input and outputs a complete plan. We then prompt the LLM to write a proof for the following theorem: \textit{Let \textbf{GP} be the generated generalized plan. If \textbf{I} is a problem instance satisfying the domain constraints, then \textbf{GP(I)} returns a plan where 1) all actions are sequentially applicable and 2) the resulting state satisfies the goal.} A valid proof for this theorem shows that the generalized plan \textbf{GP} is complete.\footnote{In our benchmarks, we only consider solvable problem instances. If the domain constraints do not guarantee solvability, then there exists no proof of the theorem.} Apart from the PDDL itself, our approach requires as input only a specification of the domain constraints, as well as a specification of state invariants. Everything else is completely automated, including the generation and debugging of \textbf{GP} as per \citet{stein2026improved}, and the proof that \textbf{GP} is complete. Figure \ref{fig:approach} illustrates our approach.

We empirically evaluate our approach on 13 commonly used benchmark domains (a subset of the domains addressed by \citet{stein2026improved}). 
Using GPT-5.6-Sol as the LLM, for 12 of these domains we obtain a generalized plan and a valid proof, showing automatically that the generalized plan is complete. 
This is a major advancement of the state of the art in automatic generalized-plan completeness proofs.

\section{Background}\label{sec:background}

\paragraph{PDDL planning.} 

In our work we focus on planning domains and problems defined in the Planning Domain Definition Language (PDDL) \cite{mcdermott20001998,DBLP:series/synthesis/2019Haslum}. In PDDL, a planning task is specified by domain $\mathcal{D}$ and a problem instance $I$ of the domain. A domain is a pair $\mathcal{D} = \langle \mathcal{P}_\mathcal{D}, \mathcal{A} _\mathcal{D}\rangle$ where $\mathcal{P}_\mathcal{D}$ defines the set of predicates for describing states and $\mathcal{A}_\mathcal{D}$ defines the set of action schemas. Each predicate $P \in \mathcal{P}_\mathcal{D}$ takes a finite number of arguments, i.e. has an arity of $k \geq 0$. 
Each action schema $A \in \mathcal{A}_\mathcal{D}$ is a triplet $\langle \text{Arg}(A), \text{Pre}(A), \text{Eff}(A) \rangle$ where $\text{Arg}(A)$ are the arguments of the action, $\text{Pre}(A)$ defines its preconditions and $\text{Eff}(A)$ its effects.

\begin{figure}[t]
\centering
\subfloat[Excerpt from the domain definition.]{\includegraphics[width=0.9\linewidth]{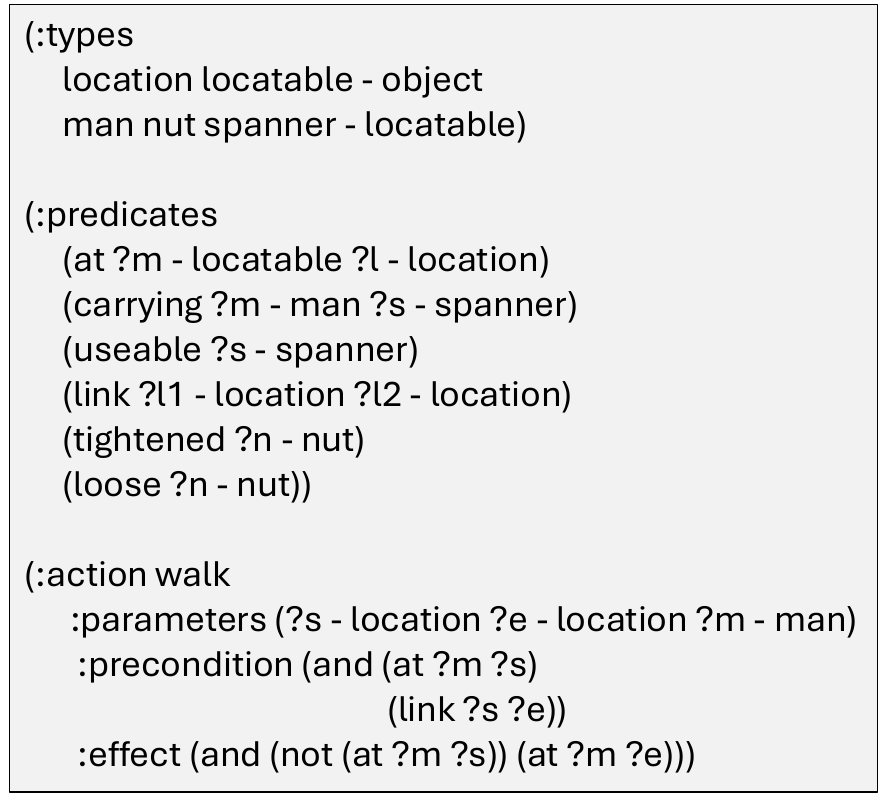}\label{fig:pddl_domain_example}}
\qquad
\subfloat[Example problem instance.]{\includegraphics[width=0.9\linewidth]{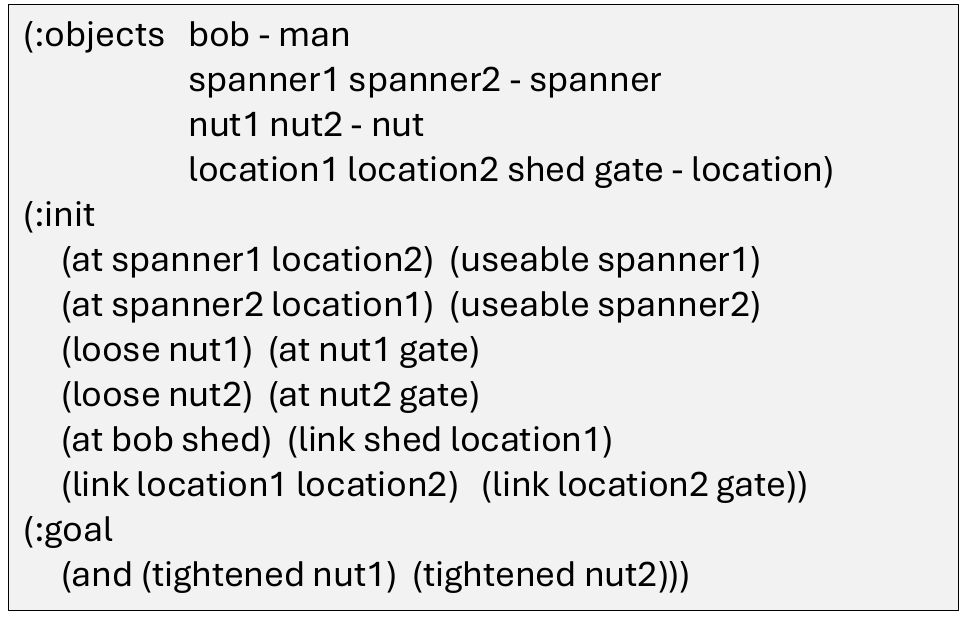}\label{fig:pddl_prob_example}}
\caption{PDDL examples from the Spanner domain.}
\label{fig:pddl_example}
\end{figure}

We focus on domains that can be expressed by a subset of PDDL that is restricted to conjunctive conditions with negation. Hence, all preconditions and effects are conjunctions of literals from $\mathcal{P}$ where all predicate arguments are also arguments of the action. We refer to the positive preconditions and effects as $\text{Pre}^{+}$ and $\text{Eff}^{+}$, and to the negative preconditions and effects as $\text{Pre}^{-}$ and $\text{Eff}^{-}$ respectively. We require $\text{Eff}^{+}(A) \cap \text{Eff}^{-}(A) = \emptyset$. 

A problem instance $I$ is a triplet $\langle O, S_0, G \rangle$ where $O$ is a finite set of objects, potentially with types, $S_0$ defines the initial state and $G$ the goal conditions. Both $S_0$ and $G$ are defined in terms of ground atoms, i.e. predicates of the domain where the arguments are replaced by objects $o \in O$, $p(o_1, ..., o_k)$. The initial state forms a set of atoms and is interpreted under the closed-world assumption, which means that all atoms that are not listed as part of the initial state are false. The goal is a conjunction of literals, $G = G^{+} \cup G^-$, that needs to be true at the end of the plan. A state $s$ satisfies a goal $G$ iff $G^{+} \subseteq s \wedge G^{-} \cap s = \emptyset$.  We write $s \models G$ if $s$ satisfies $G$. 

Figure \ref{fig:pddl_domain_example} shows an excerpt from the PDDL Spanner domain (\ref{fig:pddl_domain_example}) and Figure \ref{fig:pddl_prob_example} shows an example of a problem instance.

An action $a(o_1, ..., o_k)$ is an instance of an action schema $A$ grounded by objects from $O$. An action $a$ is applicable in a state $s$ iff all its preconditions are satisfied in $s$, i.e. $\text{Pre}^{+}(a) \subseteq s \wedge \text{Pre}^{-}(a) \cap s = \emptyset$. Applying an action $a$ in a state $s$ yields a successor state $s'$ where $s' = (s \setminus \text{Eff}^{-}(a)) \cup \text{Eff}^{+}(a)$. 
We define the successor state function $\gamma : S \times A \rightarrow S$ where S denotes the set of all states as follows: If an action $a$ is applicable in a state $s$ then $\gamma(s, a) = s' = (s \setminus \text{Eff}^{-}(a)) \cup \text{Eff}^{+}(a)$. If $a$ is not applicable then $\gamma(s, a)$ is undefined. We extend $\gamma$ to sequences of actions $[a_1, ..., a_n]$ by iterated application: $\gamma(s, []) = s$ and $\gamma(s, [a_1, ..., a_n]) = \gamma(\gamma(s, a_1), [a_2, ..., a_n])$, which is defined only if each $a_i$ is applicable in the state resulting from applying $[a_1, ..., a_{i-1}]$ to $s$. We say a plan $\pi$ is applicable from a state $s$ if $\gamma(s, \pi)$ is defined.

A plan $\pi = [a_1, ..., a_n]$ is a solution for a planning task $\langle \mathcal{D},I\rangle$ with $I = \langle O, S_0, G \rangle$ if and only if $\gamma(S_0, \pi)$ is defined and $\gamma(S_0, \pi)$ satisfies $G$.

Domains can be typed, i.e. define a hierarchy of object types. A typed domain can be converted into an equivalent untyped domain by converting types into unary predicates. 

PDDL does not provide a specification language for the set of problem instances considered. Instead, in most of the planning literature, this is encoded implicitly through a fixed set of instances considered, or as part of the program code of a problem generator associated with the domain. For example, the problem generator for the Spanner domain generates only instances where all nuts are loose in the initial state and need to be tightened in the end.
Such an implicit representation does not permit to reason about whether a generalized plan is complete w.r.t.\ a (possible infinite) set of problem instances. In our work here, we specify the instance set in terms of instance-validity constraints $C$ in Lean.

\paragraph{Generalized planning.} 
Research on generalized planning comprises a variety of different representations for generalized plans and different approaches to obtain them. 
These include policies based on features and abstractions that are hand-crafted \cite{bonet-geffner-ijcai2018} or learned \cite{bonet-et-al-aaai2019}, policies for qualitative numerical problems \cite{srivastava-et-al-aaai2011},
finite-state controllers that augment a policy with an internal memory \cite{bonet-et-al-icaps2009} and
directly synthesized programs with control flow structures such as loops and conditionals \cite[e.g.][]{Jimenez_Segovia-Aguas_Jonsson_2019, DBLP:journals/ai/SrivastavaIZ11}

More recently, approaches involving neural models have been proposed, including learning general policies with graph neural networks \cite{staahlberg2022learning, stahlberg-et-al-aaai2025} and generating programmatic generalized plans using pretrained LLMs.
\citet{Silver_2024} and \citet{stein2026improved} use LLMs to come up with strategies for solving all planning tasks from a specific domain and to implement these strategies in the form of Python functions that take a Python representation of a planning problem as input and return a sequence of actions. 
Their approaches include an automated debugging loop for the Python programs and \citet{stein2026improved} also include an automated debugging step of the strategies. 
Here, we consider generalized plans in the form of executable programs that take a problem instance as input and return a plan.

\paragraph{Completeness of generalized plans.}
Completeness guarantees for generalized plans have been approached in different ways. One line of work restricts the representation of generalized plans itself. For this restricted class of so-called one-dimensional planning problems, complete generalized plans can be found automatically using logical reasoning constraints \cite{hu-levesque-kr2010}. A complementary line of work starts from a generalized plan and determines the characterization of the set of instances that the generalized plan solves \cite{bonet-et-al-ijcai2019}, or computes the characterization of the set of covered instances jointly with the generalized plan \cite{DBLP:journals/ai/SrivastavaIZ11}.
The key concept underlying the most recent work in this direction is that of a sound abstraction: whenever the abstract generalized plan reaches the abstract goal, the resulting concrete plan is guaranteed to solve every concrete instance that is covered by the abstraction. 
\citet{cui2023automatic} show how to automatically prove the soundness of such abstractions, and hence automatically prove that the generalized plan is complete. However, this has only been shown for very simple domains (e.g. IPC Gripper) and restricted set-ups (Logistics with one vehicle) so far. 
Another approach for establishing completeness is manual analysis, which can be applied to a larger variety of domains but at the cost of manual work by experts. \citet{stein2026improved}, \citet{drexler-et-al-icaps2022} and \citet{frances-et-al-aaai2021} manually verified that their generated or hand-crafted generalized plans solve all instances of a domain. 

\paragraph{Lean.}
Lean~\cite{lean4} is a dependently typed functional programming language which enables the implementation of a generalized plan as an executable, type-checked function directly in Lean. 
Furthermore, Lean is also an interactive theorem prover. So, it is used to write programs as well as to state and prove mathematical and program-correctness theorems about them. It is based on a dependent type theory where types may depend on values. Lean is built on the Curry-Howard correspondence, thus propositions are represented as types and a proof of a proposition is a term of that type. Stating a theorem therefore amounts to writing down a type, while proving it amounts to constructing a term of that type.
A Lean representation is thus a sequence of \emph{definitions} introducing functions and datatypes, \emph{statements} expressing propositions and \emph{proofs} constructing terms that inhabit the types of their propositions. Type-checking is what constitutes verification in Lean: if the term type-checks, the statement counts as proven since a proof of a proposition is by constructing a term of the corresponding type.  Constructing such terms explicitly is impractical, so Lean offers a \emph{tactic mode} to provide such proofs as a sequence of transformation steps, such as rewriting with an equation or performing induction. Lean executes this sequence to construct the explicit proof term. This term, along with definitions and previously proven lemmas it depends on, is provided to Lean's \emph{kernel}, a small trusted program that type-checks it.

\section{Lean Representations and Correctness}\label{sec:representation}

Our objective is to obtain domain-specific generalized plans for PDDL planning domains and prove their completeness in Lean. This requires a conversion from PDDL to Lean
that faithfully preserves the semantics of the planning task. In this section, we describe our Lean representation and the conversion.

We use the subscript ``L'' to indicate that a concept is represented in Lean and no subscript for PDDL, e.g. $I$ refers to PDDL problem instance and $I_L$ to a Lean problem instance. We write \conv{\cdot} for the conversion operator that maps PDDL concepts to their Lean counterparts as defined in this section. 
\conv{I} therefore refers to the specific problem instance obtained by applying the conversion to the PDDL instance $I$. We use the inversion of the operator, i.e., \conv{\cdot}$^{-1}$, to refer to the conversion from Lean concepts back to their original PDDL counterpart. 

\subsection{Representing States and Instances}
\paragraph{States.}
Our Lean representation of a PDDL state is split into two parts: a \defined{static state} and a \defined{dynamic state} as shown in Listing~\ref{lst:spanner_lean}. 
The static and dynamic parts together comprise everything that is required for capturing all parts of a PDDL state: the set of objects, and a boolean-valued function for every predicate of the domain.

We represent objects as natural numbers $i \in \mathbb{N}$, i.e. our conversion maps each PDDL object $o\in O$ to a unique $i \in \mathbb{N}$ by enumeration, hence \conv{\cdot} is bijective on $O$. 
Each PDDL predicate $P \in \mathcal{P_D}$ of arity $k$ is represented as a boolean curried function $P_{L}:\mathbb{N}^k \rightarrow Bool$ that takes $k$ objects as input and returns a truth value. Since the predicate functions are total functions relative to the set of objects, this matches the closed-world interpretation of PDDL states.
All predicate functions for which the truth assignments cannot be changed by any action are part of the static state part together with the set of objects.
For example, the predicate \texttt{(link ?l1 - location ?l2 - location)} becomes the function \texttt{link\_p} $ : $ \Nlean $ \rightarrow $ \Nlean $ \rightarrow Bool$, and is part of the static state part because none of the action effects affect the connections between locations. All other predicate functions are part of the dynamic part. We compile away PDDL types by turning each type into a unary predicate.

Given the object mappings, a PDDL state $s$ is converted into a Lean state $s_L$ by defining the truth assignments of the predicate functions such that the following state correspondence relation holds uniquely:
\begin{definition}[State Correspondence]
\label{def:state_correspondence}
    Let $s$ be a PDDL state, $s_L$ a Lean state. $s$ and $s_L$ correspond to each other, written $s \sim s_L$, if for every predicate $P$ and tuple of objects $(o_1,...,o_k)$, $p(o_1, ..., o_k) \in s$ iff \conv{p(o_1, ..., o_k)} $= P_L(i_1,...,i_k)$ returns true, where $i_j=$\conv{o_j}.
\end{definition}

\paragraph{Instances.}
Encoding a PDDL goal condition follows a similar principle, but differs in two key aspects. 
First, predicates in the goal are defined as partial functions, as shown in Listing \ref{lst:spanner_lean}, where $Option\ Bool$ :\texttt{none} means that the atom is unconstrained, while
\texttt{some b} requires the state to assign it the truth value \texttt{b}. 
Second, the goal is defined only in terms of the predicate functions for which truth assignments can change, i.e. the dynamic part of the state. 
We refer to the goal function for a predicate $P \in \mathcal{P_D}$ as $PG_{L}$. 

\begin{figure}
\begin{lstlisting}[style=leanstyle]
structure State where
     statics : StaticState
     dynamic : DynamicState
     
structure StaticState where
     objects : List (*@\Nlean@*)
     link_p : (*@\Nlean@*) $\rightarrow$ (*@\Nlean@*) $\rightarrow$ Bool
     location_t : (*@\Nlean@*) $\rightarrow$ Bool
     locatable_t : (*@\Nlean@*) $\rightarrow$ Bool
     ...
     
structure DynamicState where
     at_p : (*@\Nlean@*) $\rightarrow$ (*@\Nlean@*) $\rightarrow$ Bool
     carrying_p : (*@\Nlean@*) $\rightarrow$ (*@\Nlean@*) $\rightarrow$ Bool
     useable_p : (*@\Nlean@*) $\rightarrow$ Bool
     ...

structure Goal where
  dynamic : GoalDynamic
  
structure GoalDynamic where
     at_p : (*@\Nlean@*) $\rightarrow$ (*@\Nlean@*) $\rightarrow$ Option Bool
     carrying_p : (*@\Nlean@*) $\rightarrow$ (*@\Nlean@*) $\rightarrow$ Option Bool
     useable_p : (*@\Nlean@*) $\rightarrow$ Option Bool
     ...
\end{lstlisting}
\captionof{listing}{Excerpt of the Lean state and goal representations of the PDDL Spanner domain (Figure \ref{fig:pddl_domain_example}).}
\label{lst:spanner_lean}
\end{figure}

A PDDL goal is converted into a Lean goal analogously to a state: \conv{G} is created such that for every predicate $P$ of arity $k$ and tuple of objects, $PG_{L}(i_1, ..., i_k) =$ true iff $p(o_1,...,o_k) \in G^{+}$ and $PG_{L}(i_1, ..., i_k) =$ false iff $p(o_1,...,o_k) \in G^{-}$ and is undefined otherwise. 
For example, if the goal specifies that nut $nut1$ needs to be tightened but does not specify the location of the man $bob$, 
and \conv{nut1} $= 3$ and \conv{bob}$= 4$ then \texttt{tightened\_p}$_G(3)$ returns true and \texttt{at\_p}$_G(4, o)$ returns none for any $o\in O$ (see Figure \ref{lst:prob_conversion} in Appendix for a full example of a converted problem instance). 

The \defined{goal satisfaction relation}, $\models_L$, is defined in Lean as a function \texttt{SatisfiesGoal : State} $\rightarrow$ \texttt{Goal} $\rightarrow$ \texttt{Prop}. \texttt{SatisfiesGoal s g} holds iff for every goal predicate and every argument, the corresponding predicate of the Lean state \texttt{s} agrees with the value that \texttt{g} requires wherever \texttt{g} requires one. We define the result as a \texttt{Prop} rather than a \texttt{Bool} because goal satisfaction appears in our completeness theorem (Theorem~\ref{theorem:gp_correct_lean}), and a theorem statement must be a proposition. Listing~\ref{lst:spanner_goal} shows an excerpt for the Spanner
domain.

\begin{figure}[t]
\begin{lstlisting}[style=leanstyle]
def SatisfiesGoal (s : State) (g : Goal) : Prop :=
  ($\forall$ var_m var_l,
    match g.dynamic.at_p var_m var_l with
    | none => True
    | some b => s.dynamic.at_p var_m var_l = b) $\wedge$
  ($\forall$ var_n,
    match g.dynamic.tightened_p var_n with
    | none => True
    | some b => s.dynamic.tightened_p var_n = b) $\wedge$
   ...
\end{lstlisting}
\captionof{listing}{Example of the Lean function checking for goal satisfaction.}
\label{lst:spanner_goal}
\end{figure}

A complete PDDL and Lean problem instance then correspond to each other if the following holds:
\begin{definition}[Instance correspondence]\label{def:inst_corresp}
    A PDDL instance $I = \langle O, S_0, G \rangle$ and a Lean instance $I_L = $\conv{I} $=\langle$\conv{O}, \conv{S_0}, \conv{G}$\rangle$ correspond to each other, written $I \sim I_L$, if $S_0 \sim$ \conv{S_0} and for any state $s$: $s \models G$ iff \conv{s} $\models_L$ \conv{G}
\end{definition}

By construction, our conversion ensures that for each PDDL instance $I$, $I \sim $\conv{I}. 

\subsection{Representing Actions and Transitions}
\paragraph{Actions.}
Actions are encoded as an inductive type \texttt{PlanAction} with one constructor per PDDL action schema  (Listing~\ref{lst:plan_action}). We use an inductive type here rather than a structure as we did for states, because a structure in Lean has a single constructor, whereas actions form a choice among several schemas. Lean requires that any
function over \texttt{PlanAction}, such as the transition function, handles every constructor. Each constructor fixes the arity 
of its schema, so that ill-formed ground actions are rejected by Lean's type checker. A Lean ground action is simply a constructor applied to all its arguments, e.g.\ \texttt{walk}$(1,2,3)$. Since all ground actions thus share the type \texttt{PlanAction}, a \defined{Lean plan} $\pi_L$ is of type \texttt{List PlanAction}.

\begin{figure}
\begin{lstlisting}[style=leanstyle]
inductive PlanAction where
  | walk (var_s : (*@\Nlean@*)) (var_e : (*@\Nlean@*)) (var_m : (*@\Nlean@*))
  | pickup_spanner (var_l : (*@\Nlean@*)) (var_s : (*@\Nlean@*)) (var_m : (*@\Nlean@*))
  | tighten_nut (var_l : (*@\Nlean@*)) (var_s : (*@\Nlean@*)) (var_m : (*@\Nlean@*)) (var_n : (*@\Nlean@*))
\end{lstlisting}
\captionof{listing}{Example of the action constructors for the action schemas from the Spanner domain.}
\label{lst:plan_action}
\end{figure}

Preconditions and effects for each action schema are represented as a pair of Lean functions. Like \texttt{SatisfiesGoal}, the precondition function for an action schema $A$ takes the arguments of the encoded action and Lean state $s_L$ to return a proposition that the grounded action is applicable in $s_L$. This function is defined as a conjunction where the conjuncts correspond to the Lean conversions of the literals in Pre$(A)$. Since state predicates are Boolean-valued, each conjunct is an equation requiring the corresponding predicate function to return \texttt{true} for a positive literal, or \texttt{false} for a negative literal. Listing \ref{lst:spanner_pre} shows an example for the preconditions of the ``walk'' action schema from the Spanner domain.

\begin{figure}
\begin{lstlisting}[style=leanstyle]
def walkPre (var_s : (*@\Nlean@*)) (var_e : (*@\Nlean@*)) (var_m : (*@\Nlean@*)) (s : State) : Prop :=
  s.dynamic.at_p var_m var_s = true $\wedge$ 
  s.statics.link_p var_s var_e = true $\wedge$ 
  s.statics.location_t var_s = true $\wedge$
  s.statics.location_t var_e = true $\wedge$ 
  s.statics.man_t var_m = true 

def walkEff (var_s : (*@\Nlean@*)) (var_e : (*@\Nlean@*)) (var_m : (*@\Nlean@*)) (s : State) : State :=
 {s with dynamic := {
  s.dynamic with
  at_p :=
    fun var_m' var_e' =>
      if var_m' = var_m $\wedge$ var_e' = var_e then
        true
      else if var_m' = var_m $\wedge$ var_e' = var_s then
        false
      else
        s.dynamic.at_p var_m' var_e'}}
\end{lstlisting}
\captionof{listing}{Example of the Lean representation of the preconditions and effects of the action schema `walk'.}
\label{lst:spanner_pre}
\end{figure}

The effect function for an action schema $A$ takes the same input as the precondition function and returns the updated state $s'_L$, in which the predicate function of each add effect in Eff($A$) returns \texttt{true} at the affected arguments, each delete effect returns \texttt{false}, and all other truth assignments are unchanged. Listing~\ref{lst:spanner_pre} shows this for the ``walk'' schema, where only \texttt{at\_p} is affected. The Lean syntax \texttt{\{s with dynamic := \ldots\}} builds a copy of \texttt{s} in which only the named field differs, and the nested \texttt{s.dynamic with \ldots} analogously replaces only \texttt{at\_p}; every other predicate function is carried over unchanged. The new \texttt{at\_p} returns \texttt{true} for the man's new location \texttt{var\_e}, \texttt{false} for the previous location \texttt{var\_s}, and the previous truth value for all other argument pairs.

\paragraph{Transitions.} By construction, for any ground action $a$ of an action schema $A$ and any state $s$ the following holds: the Lean precondition function returns true on \conv{a} and \conv{s} iff $a$ is applicable in $s$ and if so, then the effect function applied to \conv{a} and \conv{s} returns a state $s'_L$ that corresponds to the PDDL successor state $s' = \gamma(s, a)$.

\begin{figure}[t]
\centering
\begin{subfigure}[b]{0.48\linewidth}
\centering
\begin{tikzcd}[column sep=3.2em, row sep=3em]
s \arrow[r, "{\llbracket\cdot\rrbracket}", Rightarrow] \arrow[d, "{\gamma(s,a)}"] 
    & \llbracket s \rrbracket \arrow[d, "{\gamma_L(\llbracket s \rrbracket,\llbracket a\rrbracket)}"] \\
s' \arrow[r, "{\llbracket\cdot\rrbracket}", Rightarrow] 
    & \llbracket s' \rrbracket
\end{tikzcd}
\caption{Single-action transition.}
\label{fig:transitions-single}
\end{subfigure}
\hfill
\begin{subfigure}[b]{0.48\linewidth}
\centering
\begin{tikzcd}[column sep=3.2em, row sep=3em]
S_0 \arrow[r, "{\llbracket\cdot\rrbracket}", Rightarrow] \arrow[d, "{\gamma(S_0,\pi)}"] 
    & \llbracket S_0 \rrbracket \arrow[d, "{\gamma_L(\llbracket S_0 \rrbracket,\llbracket \pi\rrbracket)}"] \\
s_n \arrow[r, "{\llbracket\cdot\rrbracket}", Rightarrow] 
    & \llbracket s_n \rrbracket
\end{tikzcd}
\caption{Full-plan transition.}
\label{fig:transitions-plan}
\end{subfigure}
\caption{Conversion $\llbracket\cdot\rrbracket$ commutes with each applicable ground action when applying this single action (left). Iterating the square over a full plan shows that $\llbracket\cdot\rrbracket$ also commutes with applying a sequence of actions (right).}
\label{fig:transitions}
\end{figure}
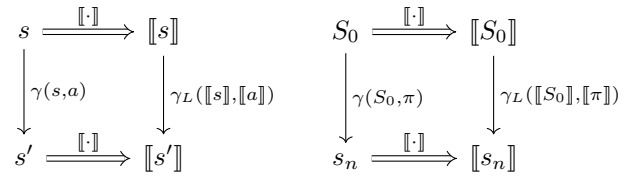

The functions defined so far are specific to a single action schema, whereas the transitions of a planning task are defined for arbitrary ground actions. We therefore combine them into \texttt{actionPre} and \texttt{actionApply} (Listing~\ref{lst:spanner_transition}), which take an arbitrary \texttt{PlanAction} and a state as input. Lean accepts these definitions only if
they match exhaustively over every constructor of \texttt{PlanAction}, so all action schemas of the domain are covered, each branch calling the precondition or effect function of its corresponding schema.
Together, \texttt{actionPre} and \texttt{actionApply} form the Lean counterpart $\gamma_L$ to the PDDL successor state function $\gamma$, with $\gamma_L(s_L, a_L) := $ \texttt{actionApply}$(a_L, s_L)$ if \texttt{actionPre}$(a_L, s_L)$ is true and undefined otherwise. Unlike $\gamma$, \texttt{actionApply} is a total function by itself, and reproduces the partial definition of $\gamma$ in combination with \texttt{actionPre}. Figure \ref{fig:transitions} (left) illustrates the following lemma: 
\begin{lemma}[Action Transitions Commute]\label{lemma:action_commute}
     For any state $s$ and ground action $a$, $\gamma(s,a)$ is defined iff $\gamma_L($\conv{s}, \conv{a}$)$ is defined. If $\gamma(s,a) = s'$ and $\gamma_L($\conv{s}, \conv{a}$) = s'_L$ then $s'_L=$\conv{s'}.
\end{lemma}

\begin{figure}
\begin{lstlisting}[style=leanstyle]
def actionPre : PlanAction $\rightarrow$ State $\rightarrow$ Prop
  | .walk var_s var_e var_m, s => walkPre var_s var_e var_m s
  | .pickup_spanner var_l var_s var_m      , s => pickup_spannerPre var_l var_s var_m s
  | .tighten_nut var_l var_s var_m var_n, s => tighten_nutPre var_l var_s var_m var_n s

def actionApply : PlanAction $\rightarrow$ State $\rightarrow$ State
  | .walk var_s var_e var_m, s => walk var_s var_e var_m s
  | .pickup_spanner var_l var_s var_m      , s => pickup_spanner var_l var_s var_m s
  | .tighten_nut var_l var_s var_m var_n, s => tighten_nut var_l var_s var_m var_n s
\end{lstlisting}
\captionof{listing}{Example of the Lean applicability and state update functions for the Spanner domain.}
\label{lst:spanner_transition}
\end{figure}

\subsection{From Single Steps to Plan Correctness}
We lift applicability and state updates from single actions to sequences by defining two recursive functions, \texttt{ValidPlan} and \texttt{runPlan} (Listing~\ref{lst:plans_lean}), the former stating the proposition under which a plan is applicable and the second computing the state it produces. The empty plan is always applicable and leaves the state unchanged. For a non-empty list, \texttt{ValidPlan} requires the first action to be applicable in the current state and the remaining actions to form a valid plan from the state that results from applying it, while \texttt{runPlan} applies the first action and continues on the remaining list. Notice that \texttt{runPlan} applies an action regardless of whether its preconditions hold and applicability is captured separately by \texttt{ValidPlan}. 
Checking for the correctness of a plan hence requires both: a plan $\pi_L$ is a solution for a problem instance with the initial state $S_{0L}$ and the goal $G_L$ iff \texttt{ValidPlan}($\pi_L, S_{0L}$) is true and \texttt{runPlan}($\pi_L, S_{0L}$) satisfies $G_L$ (see \texttt{correctSolution}, Listing \ref{lst:plans_lean}). 

\begin{figure}
\begin{lstlisting}[style=leanstyle]
def ValidPlan : List PlanAction $\rightarrow$ State $\rightarrow$ Prop
  | [], _ => True
  | a :: as, s => actionPre a s $\wedge$ ValidPlan as (actionApply a s)

def runPlan : List PlanAction $\rightarrow$ State $\rightarrow$ State
  | [], s => s
  | a :: as, s => runPlan as (actionApply a s)

def correctSolution (s : State) (g : Goal) (plan : List PlanAction) : Prop
  ValidPlan plan s $\wedge$ 
  SatisfiesGoal (runPlan plan s) g
\end{lstlisting}
\captionof{listing}{Lean functions for checking plan applicability and correctnes and for executing a plan.}
\label{lst:plans_lean}
\end{figure}

Since initial states are states and since a single action corresponds to a plan of length 1, Lemma \ref{lemma:action_commute} applies to the initial state of a problem instance and extends to plans by induction over plan length, as illustrated in Figure \ref{fig:transitions} (right).

By construction, our PDDL to Lean conversion ensures that a PDDL problem instance $I$ and its converted Lean counterpart \conv{I} 
correspond.
Therefore, a PDDL plan and its Lean counterpart agree on applicability and goal satisfaction:

\begin{lemma}[Plan Correctness Transfers to Conversion]\label{lemma:correctness}
    Let $I = \langle O, S_0, G\rangle$ be an instance of $\mathcal{D}$, and let $\pi = [a_1, ..., a_n]$ be a sequence of actions. Then $\pi$ is a correct plan solving $I$ iff \conv{\pi} is a correct plan solving the corresponding Lean planning problem \conv{I}. 
\end{lemma}

Since \conv{\cdot} is bijective on objects and each PDDL action schema corresponds to exactly one \texttt{PlanAction} constructor, \conv{\cdot} is bijective on ground action sequences and hence every Lean plan $\pi_L$ has a unique PDDL counterpart \conv{\pi_L}$^{-1}$. 

\begin{theorem}[Soundness of Lean Plan Verification]\label{theorem:correct_pddl_plan}
    Let $I = \langle O, S_0, G\rangle$ be an instance of $\mathcal{D}$. If we obtain a sequence of actions, $\pi_L:$ \textnormal{\texttt{List PlanAction}}, and applying \textnormal{\texttt{correctSolution}} to \conv{S_0}, \conv{G} and $\pi_L$ returns true then converting $\pi_L$ from Lean to PDDL gives a plan $\pi =$ \conv{\pi_L}$^{-1}$ such that $\pi$ is guaranteed to be a solution of the original PDDL problem instance $I$.
\end{theorem}

Theorem \ref{theorem:correct_pddl_plan} follows directly from Lemma \ref{lemma:correctness}: \texttt{correctSolution(\conv{S_0}, \conv{G}, $\pi_L$}) is by definition the Lean statement that $\pi_L$ is a correct plan solving \conv{I} and the bidirectionality of Lemma \ref{lemma:correctness} gives that \conv{\pi_L}$^{-1}$ is a correct plan solving $I$.

\subsection{Completeness of Generalized Plans}

We now lift the correctness from individual plans to the completeness of generalized plans. 
We define a generalized plan as a (domain-specific) Lean function \texttt{solve} that takes a Lean problem instance, separated into the initial state and the goal, as input and returns a plan in the form of a list of \texttt{PlanAction} for the input problem instance. Listing \ref{lst:solve} shows the signature of a generalized plan. 

\begin{figure}
\begin{lstlisting}[style=leanstyle]
def solve (s : State) (g : Goal) : List PlanAction :=
  -- Implementation of the generalized plan GP 
\end{lstlisting}
\captionof{listing}{Signature of the generalized plan \texttt{solve}, which takes a state and a goal and returns a plan.}
\label{lst:solve}
\end{figure}

Not every problem instance expressible in Lean needs to be handled by the \texttt{solve} function. As described in the Background, we define the set of problem instances by validity constraints $C$ on valid initial states and goals. We represent these constraints in Lean as two predicates \texttt{ValidInit} : $State \rightarrow Prop$ and \texttt{ValidGoal} : $State \rightarrow Goal \rightarrow Prop$. Completeness of a generalized plan \texttt{solve} is then defined as follows:

\begin{definition}[Generalized Plan Completeness]\label{def:completeness}
    Let $\mathcal{D}$ be a domain. A Lean function \texttt{solve} with the signature of Listing \ref{lst:solve} is a complete generalized plan for $\mathcal{D}$ if, for every initial state $s_L$ and every goal $G_L$ that satisfy \texttt{ValidInit}$(s_L)$ and \texttt{ValidGoal}$(s_L, G_L)$, the plan returned by \texttt{solve} $s_L g_L$ is applicable from $s_L$ and the resulting state satisfies $G_L$.
\end{definition}

Definition \ref{def:completeness} 
guarantees the correctness of the generated plans for an infinite number of problem instances because there are infinitely many pairs of valid initial states and goals. 
It is expressed 
as a Lean theorem of the type shown in Listing \ref{lst:solve_correct_type} which expresses exactly the universal implication statement. Listing \ref{lst:solve_correct} shows the actual Lean declaration of the theorem which can be completed by a proof candidate that can then be checked by Lean's kernel.
A Lean proof of \texttt{solveComplete} (Listing \ref{lst:solve_correct}) for a specific candidate \texttt{solve} function establishes exactly the completeness definition (Def. \ref{def:completeness}) for that candidate if accepted by Lean's kernel:
\begin{figure}
\begin{lstlisting}[style=leanstyle]
solveComplete : $\forall$ (s : State) (g : Goal),
  ValidInit s $\rightarrow$ ValidGoal s g $\rightarrow$
  ValidPlan (solve s g) s $\wedge$
  SatisfiesGoal (runPlan (solve s g) s) g
\end{lstlisting}
\captionof{listing}{Type of the completeness theorem statement for the generalized plan \texttt{solve}: for any valid initial state and goal, \texttt{solve} returns an applicable plan that reaches the goal.}
\label{lst:solve_correct_type}
\end{figure}

\begin{figure}
\begin{lstlisting}[style=leanstyle]
theorem solveComplete
   (s : State)
   (g : Goal)
   (hinit : ValidInit s)
   (hgoal : ValidGoal s g) :
   ValidPlan (solve s g) s $\wedge$
   SatisfiesGoal (runPlan (solve s g) s) g := by
  -- Proof term of the theorem 
  
\end{lstlisting}
\captionof{listing}{Lean declaration of the completeness theorem statement in Listing \ref{lst:solve_correct_type}.}
\label{lst:solve_correct}
\end{figure}

\begin{theorem}[Completeness GP Lean]\label{theorem:gp_correct_lean}
    Let $\mathcal{D}$ be a domain and let $\text{GP}_L$ be a candidate generalized plan for $\mathcal{D}$, i.e. a Lean function with the signature shown in Listing \ref{lst:solve}. 
    If there exists a proof for \texttt{solveComplete} (Def. \ref{def:completeness}) with \texttt{solve}$:= \text{GP}_L$, that is accepted by Lean's kernel, then $\text{GP}_L$ is a complete generalized plan for $\mathcal{D}$: for every problem instance $I_L$ satisfying the domain constraints, applying $GP_L$ to $I_L$ returns a plan that is applicable from the initial state and reaches the goal.
\end{theorem}

Theorem \ref{theorem:gp_correct_lean} itself requires no separate proof beyond Lean's own soundness guarantee that whenever the kernel accepts a proof term of a proposition then the proposition holds. 
Theorem \ref{theorem:gp_correct_lean} is, however, about the correctness of a generalized plan $GP_L$ with respect to Lean problem instances and not the original PDDL. We extend the correctness theorem in order to carry over to PDDL based on Lemma \ref{lemma:correctness} and Theorem \ref{theorem:correct_pddl_plan}. From Lemma \ref{lemma:correctness} and Theorem \ref{theorem:correct_pddl_plan} we get that, given our conversion \conv{\cdot}, every PDDL instance $I$ of a domain has a corresponding Lean instance, and every Lean plan that solves the converted problem instance has a PDDL counterpart that solves the original PDDL instance. Therefore, correctness of \texttt{solve} in Lean transfers to a correctness guarantee for the original PDDL. 

\begin{corollary}[Completeness of GP for PDDL Domain]\label{corr:completeness}
    Let $GP_L$ be a complete generalized plan in Lean for a domain $\mathcal{D}$ (Theorem \ref{theorem:gp_correct_lean}), and let \conv{\cdot} be our PDDL to Lean conversion. Then for every $I=\langle O, S_0, G\rangle \in \mathcal{I_D}$, GP$_L$(\conv{I}) returns a plan $\pi_L$ such that $\pi_L =$ \conv{\pi} for some plan $\pi$ that solves $I$.  
\end{corollary}

\subsection{Validity Constraints} 

The completeness theorem in Listing \ref{lst:solve_correct} is proven relative to \texttt{ValidInit} and \texttt{ValidGoal}: these two Lean predicates specify which problem instances a generalized plan is required to solve by defining the constraints on valid problem instances, which we call validity constraints. 
Our assumption here is that the user provides this specification along with the PDDL. This makes sense as the user wants a generalized plan, and the set of instances which that plan should solve is naturally part of the problem specification. In other words, we view generalized planning as the problem of plan generation replacing a single PDDL problem instance with a set of PDDL problem instances as captured by the validity constraints. (A trivial specification would admit all initial states and goals, but this rarely makes sense; consider e.g.\ Blocksworld initial states with cyclic towers.)

Listing \ref{lst:constraints} shows examples of validity constraints for the Spanner domain. The goal constraints take the goal and the initial state as parameters, hence allowing dependencies between initial state and goal.

\begin{figure}[t]
\begin{lstlisting}[style=leanstyle]
def AtLeastAsManySpannersAsNuts (s : StaticState) : Prop :=
  (s.objects.filter (fun x => s.nut_t x)).length $\leq$ (s.objects.filter (fun x => s.spanner_t x)).length

def ValidLinkParam (s : StaticState) : Prop :=
  $\forall$ var_l1 var_l2, s.link_p var_l1 var_l2 = true $\rightarrow$ s.location_t var_l1 = true $\wedge$ s.location_t var_l2 = true

def LinkAcyclic (s : StaticState) : Prop :=
  $\forall$ l, $\neg$ Relation.TransGen (fun a b => s.link_p a b = true) l l

def InitNutsLoose (s : State) : Prop :=
  $\forall$ n, s.statics.nut_t n = true $\rightarrow$ s.dynamic.loose_p n = true

def GoalAllNutsTightened (initial : State) (g : Goal) : Prop :=
  $\forall$ n, initial.statics.nut_t n = true $\rightarrow$ g.dynamic.tightened_p n = some true

\end{lstlisting}
\captionof{listing}{Examples of constraints on valid instances of the Spanner domain.}
\label{lst:constraints}
\end{figure}

Lean's logic is the calculus of inductive constructions~\cite{pfenning:mohring:mfps-90}, used for defining relations by induction and quantifying over predicates. This allows directly expressing properties such as reachability and hence the absence of cycles of any length. Thus constraints such as the location graph in Spanner forming a linear sequence are straightforward as we can define the transitive closure of \texttt{link\_p} and state that no location is reachable from itself (\texttt{LinkAcyclic}). 

The validity of an initial state (\texttt{ValidInit}) is then defined as a conjunction of the initial state constraints and validity of a state in general, which itself is again a conjunction of the constraints on the dynamic state part and the constraints on the static state part. 
Similarly, goal validity of a state (\texttt{ValidGoal}) is defined as a conjunction of the goal constraints.

\section{Generalized Planning in Lean with LLMs}\label{sec:our_form}

The previous section described the theoretical foundation of our approach and the prerequisites for representing generalized plans for PDDL planning tasks in Lean and proving their completeness. 
In this section, we now describe how we obtain the generalized plans and completeness proofs. Given a PDDL domain and the specifications of the validity constraints, our pipeline 1) automatically generates its Lean representation, 
2) lets an LLM generate a candidate generalized plan \texttt{solve}, and 3) uses again an LLM to generate the proof that \texttt{solve} satisfies the correctness theorem from Listing \ref{lst:solve_correct}. 
All these steps, including potential debugging of the LLM-generated Lean code, are fully automated.

\paragraph{Generating the Lean domain representation.}
In order to obtain the Lean formalizations, we implemented a domain-independent Python converter that takes as input the PDDL domain, and optionally a PDDL problem, and outputs the converted Lean representation, i.e. the state, action and instance representations described above (see Listings \ref{lst:spanner_lean}, \ref{lst:plan_action}-\ref{lst:spanner_transition}). 
Our implementation for processing the PDDL covers a subset of PDDL that is restricted to conjunctive conditions with negation, no action costs, and can be applied to typed and untyped domains\footnote{In order to extend the approach to a larger PDDL fragment only the PDDL processing part needs to be adapted.}. 
The converter is purely symbolic and deterministic\footnote{The only exception with respect to the determinism concerns the order of iterations, e.g. in which the actions are converted and objects are enumerated. None of the iteration orderings, however, has any effect on the validity of our results.}, and applies the conversion described in the previous section in the same structured way to any domain of the covered PDDL fragment. Therefore, all claims and guarantees that depend on the specific conversion mechanism described before hold for the Lean representation of any domain obtained from our converter.

\paragraph{Generating the generalized plan in Lean}
To generate generalized plans with LLMs, we follow the process introduced by \citet{stein2026improved}, which consists of two main steps: 1) deriving the pseudocode strategy for solving a planning domain, and 2) implementing that strategy in a programming language. As our focus here is on completeness proofs rather than on finding strategies, we start from the pseudocode strategies of \citet{stein2026improved}, check them and revise them if necessary. 
We provide an LLM with the revised strategies as well as the complete Lean representation of the target domain and have it implement the \texttt{solve} function (Listing~\ref{lst:solve}) such that \texttt{correctSolution} (Listing~\ref{lst:plans_lean}) holds. This code is debugged in two steps. Statically, Lean's kernel must accept the code. Dynamically, the candidate generalized plan is run on a set of debugging tasks and the plans it returns are converted to PDDL and validated with VAL~\cite{DBLP:conf/ictai/HoweyLF04}. 
If either step fails, the LLM receives feedback about the specific error and is prompted to revise its generated code.
If the generated candidate generalized plan returns correct solutions for all debugging tasks within a defined number of iterations, the process moves on to the proof step. An example of a generalized plan generated for the Spanner domain can be found in Listing \ref{lst:example_gp}.

\paragraph{Generating the completeness proof.}

Once a generalized plan has been generated, we ask the LLM to prove that it is correct, i.e., to state and prove the correctness theorem shown in Listing~\ref{lst:solve_correct}. As input, the LLM receives (i)~the complete Lean code from the previous step, consisting of the Lean representation and the LLM-generated generalized plan, (ii) lemmas stating that actions preserve state validity, described in the next section, and (iii)~helper lemmas describing which parts of a state an action changes and which it leaves unchanged. These helper lemmas are generated automatically from the PDDL by a symbolic, domain-independent generator (see Listing \ref{lst:add_lemmas_spanner} and \ref{lst:add_lemmas_domain_independent} in Appendix). 

We again use an automatic debugging procedure to let the LLM revise its generated proof in case the generated proof is rejected. If Lean reports any error, or the proof fails one of the checks against cheating described below, the error message is passed back to the LLM, together with a prompt asking it to regenerate the complete proof.

Depending on the specific planning domain and the concrete implementation of the generalized plan, completeness can be hard to prove and the simple, basic approach described may not work.  
In particular, we observe that in these cases the LLM explicitly replies that proving completeness is too complex. To address this we propose a second, iterative approach inspired by \citet{ospanov2025apollo}, which, as our experiments show, works better for those complex cases. The LLM is given the same information as in the basic approach but is instructed to provide a proof sketch that states all lemmas it requires. Notably, this approach allows the LLM to decide which lemmas it proves immediately in its first response and which are marked unproven for now. The remaining lemmas are then proven in subsequent iterations. The completeness proof is only accepted once it depends on no unproven lemmas.

To debug the proof sketch, we split it into its individual declarations, i.e.\ definitions, lemmas and theorems. These declarations are added to the Lean representation one at a
time, checking the code after each addition. When a declaration is rejected, the LLM revises the declaration based on the code accepted so far and the error message. Thus, errors are repaired locally without discarding the parts Lean has already accepted in contrast to the basic approach where the complete proof needs to be regenerated. The process ends when all declarations are accepted or the maximum number of iterations is reached. During the debugging, to account for declarations missing in the initial sketch, the LLM is explicitly allowed to generate more declarations than the one that failed. They are then inserted into the proof sketch at the same place as the declaration that is currently being debugged.

\paragraph{Preventing LLM cheating.}
LLMs generating Lean code are known to find workarounds that make the code compile without establishing the intended result~\cite{cheating:arxiv-26, google:swarm-cheat:arxiv-26}, and recent work analyzing Lean benchmarks has found related evaluation-time loopholes that can affect the reliability of evaluation results \cite{lean-bug:arxiv-26}. 
There are three ways in which this can happen in our setting. First, the LLM may change \texttt{correctSolution} either by directly editing the statement itself, e.g. adding hypotheses that make the statement easier to prove, or by altering the semantics, e.g. by redefining notation or introducing new definitions~\cite{google:swarm-cheat:arxiv-26}. Second, the proof may silently rely on unjustified axioms which do not appear in the theorem's type. Third, a proof may be accepted while bypassing Lean's kernel checks due to metaprogramming or bugs in incremental compilation within Lean.

We address these with four checks. We reject code that alters how subsequent code is parsed or elaborated, such as custom notation or macros. We further verify that the final code contains a theorem exactly as written in Listing~\ref{lst:solve_correct_type}. Together, these rule out the first form of cheating. For the second form of cheating, we check that the proof does not depend on axioms beyond the standard ones.\footnote{\url{https://lean-lang.org/doc/reference/latest/Axioms/#standard-axioms}}
Finally, for efficiency, our iterative debugging loop is performed incrementally in REPL.\footnote{\url{https://github.com/leanprover-community/repl}}
As a sanity check, the complete code is rebuilt from scratch as a self-contained module and rechecked by Lean's standalone checker, thus ruling out the third form of cheating as well.

\paragraph{Invariants.}
In addition to the types of constraints described above, we specify a set of state invariants: properties that hold in the initial state and remain true in any state reachable from it, such as \texttt{NutTightenedLooseExhaustive} shown in Listing \ref{lst:invariant} which specifies that each nut is always either tightened or loose. Every invariant of this kind is a consequence of the initial state constraints together with the action schemas. 
In practice, however, we provide these invariants explicitly as part of the valid state specification because doing so removes the burden of re-deriving them from scratch from the LLM. At the same time, adding them to the constraints does not add a substantial additional amount of work. For example, without invariants, the initial state constraints need to include both a constraint that all nuts are loose (\texttt{InitNutsLoose}) and one specifying that no nut is tightened. When adding \texttt{NutTightenedLooseExhaustive} as an invariant constraint, the constraint about all nuts being not tightened becomes redundant. 

\begin{figure}
\begin{lstlisting}[style=leanstyle]
def NutTightenedLooseExhaustive (s : State) : Prop :=
  $\forall$ n, s.statics.nut_t n = true 
    $\rightarrow$ s.dynamic.tightened_p n = true $\vee$ 
       s.dynamic.loose_p n = true
\end{lstlisting}
\captionof{listing}{Examples of a state invariant constraint of the Spanner domain.}
\label{lst:invariant}
\end{figure}

%

We define a \texttt{ValidState} as a conjunction over all constraints that apply to arbitrary states (i.e. not only initial states). We then automatically generate one lemma per action schema for proving that applying the specific action in a valid state yields again a valid state. Then, an LLM is prompted to generate the proof.
We include an automatic debugging loop where the LLM has up to five tries for generating these proofs. Here, the LLM is prompted to generate, and debug, the proofs for all actions together instead of for each action individually.

\section{Experiments}
\subsection{Pipeline and Set-up}

We test our approach using GPT-5.6-Sol with temperature set to 1 and reasoning set to high. We select 13 of the 17 domains on which also \citet{stein2026improved} ran their experiments. We exclude Gripper because of its similarity to Grippers, Beluga because LLMs struggle to find generalized plans there \cite{stein2026improved} and Minigrid and Visitall as the specification of solvability is not straightforward\footnote{\citet{stein2026improved} used a version of Visitall that allows missing cells in the grid}. For all the domains that we test our approach on except Rovers, \citet{stein2026improved} manually proved that their generalized plans were complete.
We define the domain constraints on valid instances based on the problem generators, example problems from datasets, potential domain descriptions, and the PDDL domain definition.
For each domain, we select six small problem instances from existing datasets as debugging tasks (see Table \ref{tab:data_sources} in Appendix for sources of datasets). 

\subsection{Generalized Plans and Proofs}
\paragraph{Results generalized plans.}
The approach for generating the generalized plan (GP) described above is applied for obtaining the GPs for all 13 domains. We set the maximum number of debugging iterations to four. Table \ref{tab:results} presents the results. For all 13 domains, the LLM successfully generated generalized plans that were accepted by Lean and returned plans that were accepted by VAL for all six debugging tasks. The number of interactions (N interactions) refers to the number of input-output pairs of the LLM generation process for the generalized plan (GP) and the proof (Prf.), and includes both the initial versions generated by the LLM as well as the debugging iterations. 
For 10 of the domains, the first generated version of the generalized plan already solved the debugging tasks, and for the other three domains a single debugging iteration was enough. Given that we made sure to already provide the LLM a correct strategy for solving the domain, a low number of debugging iterations is expected here. 

\paragraph{Results proofs.}
As described above, we consider two different approaches for generating the completeness proofs: a basic approach where the complete proof is generated and debugged as a whole and an iterative approach where the LLM first generates a sketch of the proof and then debugs the individual declarations (e.g., lemmas, definitions, theorems, etc.). 

We first tested the basic approach for all domains with the maximum iterations for debugging set to six. For 7 of the 13 domains, the LLM successfully completed the completeness statement from Listing \ref{lst:solve_correct} with a proof that was accepted by Lean's kernel. The number of generated proof versions (i.e. initial + debugging iterations) and the time required for obtaining the generalized plans and proofs for these 7 domains are shown in the upper part of Table \ref{tab:results}.

For the other six domains, the LLM was not able to generate the completeness proof using the basic set-up. In particular, for all of these domains the LLM explicitly stated in the first response that providing a complete proof at once is not possible due to its complexity and the number of required additional lemmas. We therefore ran the iterative approach for those six domains. As the debugging is applied at the level of individual declarations here, we set the debugging limit such that the debugging terminates if the same declaration (as identified by the name of e.g. the lemma) is debugged six times in a row without success. Using the iterative approach, we successfully obtain proofs accepted by Lean's kernel for all domains except Transport. 
The lower part of Table \ref{tab:results} shows the maximum number of iterations required for the same declaration (Decl.) as well as the total number of interactions (Prf.) for obtaining the complete correct proof. 

Note, however, that the lack of a proof for Transport does not mean that the generalized plan does not solve all instances of the domain, but that the LLM was not capable of generating one. In order to still get an idea of how well the generalized plan performs, we tested it on the easy and medium test splits of the IPC Learning track 2023 \cite{taitler-et-al-aimag2024} on which it achieved perfect coverage. 

\begin{table}[ht]
    \centering
    \begin{tabular}{l|r|r|r|r|r|r}
    \hline
        \multirow{2}{*}{Domain} & \multicolumn{3}{c|}{N interactions} & \multicolumn{3}{c}{Time}  \\
         & GP & Prf. & Decl. & GP & Prf. & total\\
         \hline\hline
         \multicolumn{7}{l}{Basic proof generation} \\
         \hline
         Delivery & 1 & 3 &-- & 121 & 671 & 792 \\
         Ferry & 2 & 3 &-- & 104 & 586 & 690 \\
         Grippers & 2 & 3 & --& 119 & 661 & 780 \\
         Heavy & 1 & 3 & --& 36 & 495 & 531 \\
         Hiking & 1 & 2 & --& 113 & 582 & 694 \\
         Logistics & 1 & 3 & --& 157 & 1025 & 1182\\
         Satellite & 1 & 4 &-- & 224 & 1383 & 1607 \\
         \hline\hline
         \multicolumn{7}{l}{Iterative proof generation}\\
         \hline
         Blocks. & 2 & 24 & 3 & 315 & 1505 & 1819\\
         Goldminer & 1 & 36 & 2 & 290 & 2524 & 2814\\
         Miconic & 1 & 11 & 2 & 204 & 1302 & 1506\\
         Rovers & 1 & 45 & 3 & 307 & 2914 & 3221 \\
         Spanner & 1 & 36 & 4 & 214 & 2735 & 2950\\
         Transport & 1 & -- & -- & 239 & -- & --\\
         \hline
    \end{tabular}
    \caption{Number of LLM interactions and run time of the pipeline (in seconds) for generating the generalized plan (GP) and its completeness proof (Prf.). ``N interactions'' counts the number of interactions, i.e. input-output pairs, with the LLM, including the initial generation and all debugging iterations. The upper part shows the domains for which the basic approach succeeded, the lower part shows the results using the iterative approach on the other domains. For the iterative approach, ``Decl.'' reports the maximum number of debugging iterations required for a single declaration.}
    \label{tab:results}
\end{table}

\paragraph{Generation times.} 
Table \ref{tab:results} also shows how long it takes to obtain the generalized plans and the completeness proofs. The required time varies a lot between the different domains, ranging from 9 minutes for Heavy to almost one hour (54 minutes) for Rovers. 
Generating the generalized plans takes significantly less time even if the number of iterations is only slightly higher. As shown in Table \ref{tab:n_chunks}, the Lean code for the proofs consists also of significantly more individual declarations, both with respect to the number of declarations the LLM generates and the number of declarations already provided to the LLM in the prompt. The number of declarations provided to the LLM is necessarily larger for the proof generation because it includes the complete Lean code (provided and LLM-generated) for the generalized plan. 

\begin{table}[ht]
    \centering
    \begin{tabular}{l|r|r|r|r}
    \hline
         \multirow{2}{*}{Domain} & \multicolumn{2}{c|}{N decl. GP} & \multicolumn{2}{c}{N decl. Prf.} \\
         & given & LLM & given & LLM \\
         \hline\hline
         \multicolumn{5}{l}{Basic proof generation} \\
         \hline
         Delivery & 56 & 3 & 92 & 15 \\
         Ferry & 45 & 7 & 83 & 7 \\
         Grippers & 51 & 9 & 102 & 23 \\
         Heavy & 47 & 5 & 76 & 9\\
         Hiking & 45 & 5 & 65 & 16\\
         Logistics & 61 & 16 & 133 & 46 \\
         Satellite & 56 & 10 & 116 & 17\\
         \hline\hline
         \multicolumn{5}{l}{Iterative proof generation}\\
         \hline
         Blocks. & 52 & 10 & 122 & 42 \\
         Goldminer & 78 & 16 & 194 & 71 \\
         Miconic & 55 & 7 & 103 & 27 \\
         Rovers & 120 & 17 & 293 & 66 \\
         Spanner & 60 & 9 & 104 & 46 \\
         \hline
    \end{tabular}
    \caption{Number of Lean code declarations (i.e., lemmas, theorems, and definitions) involved in generating the generalized plan (GP) and its completeness proof (Prf.). ``Given'' reports the number of declarations provided to the LLM as part of the prompt (e.g., the domain formalization and, for proof generation, the complete GP code); ``LLM'' reports the number of individual declarations generated by the LLM itself.}
    \label{tab:n_chunks}
\end{table}

\section{Related Work}

\paragraph{Software verification.}
Verifying the correctness of synthesized programs has long been studied in software verification and program synthesis. Classical work established that a verifier can be used not only to check a synthesized program in the end, but to guide synthesis itself, iteratively refining candidate programs against counterexamples until a verified implementation is found \cite{10.1145/1168857.1168907}. 
More recently, this combination of synthesis and verification has been extended to use LLMs as the source of the candidate programs. \citet{bhatia2024} use an LLM to jointly generate a program and a proof of its functional equivalence to a source program, translating both into an automated theorem prover to verify correctness for all program states rather than testing on a finite test set. 
\citet{aggarwal2025} similarly address formally verified code generation by iteratively translating and refining candidate programs against feedback from a formal verifier, and \citet{chenICLR} fine-tune models on synthesized proof data for proof generation and debugging in Rust. 

In all these approaches, an external, trusted verifier provides the correctness guarantee for an LLM-generated program, e.g. a transpiled function \cite{bhatia2024} or a Verus-annotated Rust function \cite{aggarwal2025}, but the guarantees are only checked once against one specification. 
We adopt the same overall pattern here, but for programs that must remain correct across every valid instance of a planning domain, not a single specific input-output specification.

\paragraph{LLMs for theorem proving.}
\citet{baldur23} introduce an approach to  automatically generate and repair complete proofs using LLMs based on feedback from the Isabelle proof assistant to iteratively debug incorrect generated proofs. They fine-tune models and evaluate them on a benchmark of general mathematical theorems rather than software or planning related theorems. \citet{leancopilots} integrate LLM-based tactic suggestions, proof search and premise selection directly into the Lean workflow, focusing on interactively developing proofs for math. 
Most closely related to our proof-debugging approach is the work by \citet{ospanov2025apollo}: they isolate failing sub-lemmas in an LLM-generated proof sketch and repair them individually, using a hybrid approach that combines automated solvers and LLM calls, instead of debugging the whole proof at once. Their focus is on competition-style mathematical benchmarks.

\paragraph{Formally verified planning components.}
In ensuring the correct PDDL semantics in Lean, our work loosely relates to works on formal verification of components or aspects of planning. Previous work has worked on formalizing classical planning domains for the purpose of verification. One line of work formalizes the semantics of a PDDL fragment directly inside an interactive theorem prover: \citet{abdulaziz} formalize STRIPS-like PDDL in Isabelle/HOL and derive a formally verified plan validator, and follow-up work extends this to a verified compositional planning algorithm \cite{abdulaziz_et_al:LIPIcs:2019:11059} and a formally verified SAT-based planner \cite{abdulaziz-kurz-aaai2023}. 
Another line of work lets planners emit independently checkable certificates alongside their output, without formalizing the planner itself: \citet{eriksson-et-al-icaps2017} introduce certificates of unsolvability that a planner-independent validator can verify, and \citet{mugdan-et-al-icaps2023} extend this approach of certifying algorithms to certificates of plan optimality. 
Like our work, both lines of work target machine-checked guarantees in the context of classical planning but verify a fixed planning algorithm's output, e.g. single plans, (un)solvability claims, rather than verifying correctness across the entire set of domain instances.

\section{Conclusion}

We introduce an approach for automatically obtaining generalized plans together with machine-checked completeness proofs, using LLMs to generate both of them directly in Lean. 
%
%
Out of 13 commonly used benchmark domains, our approach produced provably complete generalized plans in 12. To our knowledge, this is the first time that generalized plans were automatically proved complete at such a scale, with prior work limited to simple domains and/or instance sets. This is a major advancement of the state of the art in automatic generalized-plan completeness proofs.



\section*{Acknowledgments}
The work was partially funded by the Deutsche Forschungsgemeinschaft (DFG, German Research Foundation) under the project number 232722074 – SFB 1102. It was also funded in part by the Deutsche Forschungsgemeinschaft -- GRK 2853/1 “Neuroexplicit Models of Language, Vision, and Action” - project number 471607914. We thank Jan Reineke, Dan Fi\v{s}er and Nicola M\"uller for insightful discussions.

\bibliography{aaai2027}
\appendix

\section{Datasets}

Table \ref{tab:data_sources} lists the datasets from which the debugging problem instances, as well as the instance generators based on which the validity constraint specifications were created.

\begin{table*}[ht]
    \centering
    \begin{tabular}{l|l|l}
    \hline
         \textbf{Domain} & \textbf{Source of debugging problems} & \textbf{Instance Generator}\\
         \hline\hline
         delivery & \multicolumn{2}{c}{\citet{Silver_2024}}  \\

         ferry & \multicolumn{2}{c}{IPC Learning track 2023 \cite{taitler-et-al-aimag2024}} \\
         heavy & \multicolumn{2}{c}{\citet{Silver_2024}} \\

         hiking & \multicolumn{2}{c}{\citet{Silver_2024}}  \\

         miconic & \multicolumn{2}{c}{\citet{Silver_2024}}  \\

         spanner& \multicolumn{2}{c}{IPC Learning track 2023 \cite{taitler-et-al-aimag2024}} \\
         blocksworld & \multicolumn{2}{c}{IPC Learning track 2023 \cite{taitler-et-al-aimag2024}} \\

         goldminer & \citet{stein2026improved} & \citet{seipp-et-al-zenodo2022}\\
  
         grippers & \citet{stein2026improved} & \citet{seipp-et-al-zenodo2022}\\

        logistics & \citet{stein2026improved} & \citet{seipp-et-al-zenodo2022}\\
  
         rovers & \multicolumn{2}{c}{IPC Learning track 2023 \cite{taitler-et-al-aimag2024}} \\

         satellite & \multicolumn{2}{c}{IPC Learning track 2023 \cite{taitler-et-al-aimag2024}} \\

         transport & \citet{stein2026improved} & \citet{seipp-et-al-zenodo2022}\\
         \hline
    \end{tabular}
    \caption{The origin of the debugging tasks that we used for our experiments and the instance generators based on which we created the validity constraints.}
    \label{tab:data_sources}
\end{table*}

\section{Additional Representation Examples}

Figure \ref{lst:prob_conversion} shows a side-by-side example of a PDDL problem instance with two spanners, two nuts and four locations (including shed and gate) (left part) and the corresponding Lean problem instance (right part). 

Listing \ref{lst:add_lemmas_spanner} shows examples of the additional, domain-specific lemmas about how states change (and do not change) that are generated symbolically based on the PDDL and provided as helper lemmas to the LLM. Listing \ref{lst:add_lemmas_domain_independent} shows additional symbolically generated lemmas that the LLM is provided for an domain. 

\begin{figure*}[ht]
\begin{minipage}{0.26\textwidth}
\begin{lstlisting}[style=pddlstyle]
(define (problem prob0)
 (:domain spanner)
 (:objects 
   bob - man
   spanner1 - spanner
   spanner2 - spanner
   nut1 - nut
   nut2 - nut
   loc1 - location
   loc2 - location
   shed - location
   gate - location
 )
 (:init 
   (at bob shed)
   (at spanner1 loc2)
   (useable spanner1)
   (at spanner2 loc1)
   (useable spanner2)
   (loose nut1)
   (at nut1 gate)
   (loose nut2)
   (at nut2 gate)
   (link shed loc1)
   (link loc1 loc2)
   (link loc2 gate)
 )
 (:goal
  (and
   (tightened nut1)
   (tightened nut2)
  )
 )
)

\end{lstlisting}
\end{minipage}
\hfill
\begin{minipage}{0.72\textwidth}
\begin{lstlisting}[style=leanstyle]
def problemStatics : StaticState :=
{
  objects := [0, 1, 2, 3, 4, 5, 6, 7, 8]
  link_p := fun x0 x1 => (x0 == 5 && x1 == 6) || (x0 == 6 && x1 == 8) || (x0 == 7 && x1 == 5),
  location_t := fun x0 => (x0 == 5) || (x0 == 6) || (x0 == 7) || (x0 == 8),
  locatable_t := fun x0 => (x0 == 0) || (x0 == 1) || (x0 == 2) || (x0 == 3) || (x0 == 4),
  man_t := fun x0 => (x0 == 0),
  nut_t := fun x0 => (x0 == 3) || (x0 == 4),
  spanner_t := fun x0 => (x0 == 1) || (x0 == 2)
}

def problemDynamic : DynamicState :=
{
  at_p := fun x0 x1 => (x0 == 0 && x1 == 7) || (x0 == 1 && x1 == 6) || (x0 == 2 && x1 == 5) || (x0 == 3 && x1 == 8) || (x0 == 4 && x1 == 8),
  carrying_p := fun x0 x1 => false,
  useable_p := fun x0 => (x0 == 1) || (x0 == 2),
  tightened_p := fun x0 => false,
  loose_p := fun x0 => (x0 == 3) || (x0 == 4)
}

def goalDynamic : GoalDynamic :=
{
  at_p := fun x0 x1 => none,
  carrying_p := fun x0 x1 => none,
  useable_p := fun x0 => none,
  tightened_p := fun x0 => if x0 == 3 then some true else if x0 == 4 then some true else none,
  loose_p := fun x0 => none
}}
\end{lstlisting}
\end{minipage}
\caption{Example of a PDDL definition of a problem instance from the Spanner domain (left) and the corresponding, converted Lean problem instance (right).}
\label{lst:prob_conversion}
\end{figure*}

\begin{figure*}[t]
\begin{lstlisting}[style=leanstyle]
-- walk does not change any predicate function from the static state part
lemma walk_statics (var_start var_end var_m : Obj) (s : State) :
    (walk var_start var_end var_m s).statics = s.statics := rfl

-- walk changes at_p for the man walking and the end location to true
lemma walk_at_p_eq1 (var_start var_end var_m : Obj) (s : State) :
    (walk var_start var_end var_m s).dynamic.at_p var_m var_end = true := by
  unfold walk
  simp

-- walk never changes at_p for locations other than the start and end location parameters
lemma walk_at_p_ne_var_l (var_start var_end var_m : Obj) (s : State) {var_end' : Obj} (h1 : var_end' $\neq$ var_end) (h2 : var_end' $\neq$ var_start) :
    (walk var_start var_end var_m s).dynamic.at_p var_m var_end' = s.dynamic.at_p var_m var_end' := by
  unfold walk
  simp [h1, h2]

-- walk never changes carrying_p
lemma walk_carrying_p (var_start var_end var_m : Obj) (s : State) :
    (walk var_start var_end var_m s).dynamic.carrying_p = s.dynamic.carrying_p := rfl

-- Applying any plan does not change any predicate functions from the static state part 
lemma runPlan_statics (plan : List PlanAction) (s : State) :
    (runPlan plan s).statics = s.statics := by
  induction plan generalizing s with
  | nil => simp [runPlan]
  | cons a as ih =>
      simp only [runPlan]
      rw [ih]
      cases a with
      | walk var_s var_e var_m => exact walk_statics var_s var_end var_m s
      | pickup_spanner var_l var_s var_m => exact pickup_spanner_statics var_l var_s var_m s
      | tighten_nut var_l var_s var_m var_n => exact tighten_nut_statics var_l var_s var_m var_n s

-- Applying an action in a valid state yields a valid successor state
lemma action_preserves_wf
    {a : PlanAction}
    {s : State}
    (hwf : WellFormed s)
    (hpre : actionPre a s) :
    WellFormed (actionApply a s) := by
  cases a with
  | walk var_start var_end var_m =>
      exact walk_preserves_wf var_start var_end var_m s hwf hpre
  | pickup_spanner var_l var_s var_m =>
      exact pickup_spanner_preserves_wf var_l var_s var_m s hwf hpre
  | tighten_nut var_l var_s var_m var_n =>
      exact tighten_nut_preserves_wf var_l var_s var_m var_n s hwf hpre

\end{lstlisting}
\captionof{listing}{Examples of the additional, symbolically generated, lemmas provided the LLM for generating the completeness proof for the Spanner domain.}
\label{lst:add_lemmas_spanner}
\end{figure*}

\begin{figure*}[t]
\begin{lstlisting}[style=leanstyle]
-- Applying a valid plan in a valid state yields a valid state
lemma validPlan_preserves_wf
    {plan : List PlanAction}
    {s : State}
    (hplan : ValidPlan plan s)
    (hwf : WellFormed s) :
    WellFormed (runPlan plan s) := by
  induction plan generalizing s with
  | nil =>
      simpa [runPlan]
  | cons a as ih =>
      simp only [ValidPlan] at hplan
      rcases hplan with $\langle$ha, has$\rangle$
      have hwf' :=
        action_preserves_wf (s := s) hwf ha
      exact ih has hwf'

-- Additional lemmas for combining sub plans
lemma runPlan_append (as bs : List PlanAction) (s : State) :
    runPlan (as ++ bs) s = runPlan bs (runPlan as s) := by
  induction as generalizing s with
  | nil => simp [runPlan]
  | cons a as ih => simp [runPlan, ih]

lemma validPlan_append (as bs : List PlanAction) (s : State) :
    ValidPlan (as ++ bs) s $\leftrightarrow$ ValidPlan as s $\wedge$ ValidPlan bs (runPlan as s) := by
  induction as generalizing s with
  | nil => simp [ValidPlan, runPlan]
  | cons a as ih => simp [ValidPlan, runPlan, ih, and_assoc]
\end{lstlisting}
\captionof{listing}{The additional, symbolically generated, lemmas provided the LLM for generating the completeness proof for any domain.}
\label{lst:add_lemmas_domain_independent}
\end{figure*}

\onecolumn
\setcounter{lstlisting}{13}
\begin{lstlisting}[style=leanstyle2, caption={Example of an LLM-generated generalized plan for the Spanner domain, consisting of the main \texttt{solve} function and helper declarations.}, label={lst:example_gp}]
structure GPBuild where
  actions  : List PlanAction
  state    : State
  location : Obj

def gpFindMan (s : State) : Obj :=
  match s.statics.objects.find? (fun o => s.statics.man_t o) with
  | some m => m
  | none   => 0

def gpFindManLocation (s : State) (m : Obj) : Obj :=
  match s.statics.objects.find? (fun l =>
    s.statics.location_t l && s.dynamic.at_p m l) with
  | some l => l
  | none   => 0

def gpFindNextLocation (s : State) (l : Obj) : Option Obj :=
  s.statics.objects.find? (fun next =>
    s.statics.link_p l next)

def gpFindCarriedUseableSpanner (s : State) (m : Obj) : Option Obj :=
  s.statics.objects.find? (fun sp =>
    s.statics.spanner_t sp &&
    s.dynamic.carrying_p m sp &&
    s.dynamic.useable_p sp)

def gpCollectAt : List Obj $\rightarrow$ Obj $\rightarrow$ Obj $\rightarrow$ State $\rightarrow$ GPBuild
  | [], l, _, s =>
      {
        actions := []
        state := s
        location := l
      }
  | sp :: rest, l, m, s =>
      if s.statics.spanner_t sp = true $\wedge$
         s.dynamic.at_p sp l = true $\wedge$
         s.dynamic.useable_p sp = true then
        let a := PlanAction.pickup_spanner l sp m
        let s' := actionApply a s
        let result := gpCollectAt rest l m s'
        {
          actions := a :: result.actions
          state := result.state
          location := l
        }
      else
        gpCollectAt rest l m s

def gpTraverse : Nat $\rightarrow$ Obj $\rightarrow$ Obj $\rightarrow$ State $\rightarrow$ GPBuild
  | 0, l, m, s =>
      gpCollectAt s.statics.objects l m s
  | Nat.succ fuel, l, m, s =>
      let collected := gpCollectAt s.statics.objects l m s
      match gpFindNextLocation collected.state l with
      | none =>
          collected
      | some next =>
          let a := PlanAction.walk l next m
          let s' := actionApply a collected.state
          let result := gpTraverse fuel next m s'
          {
            actions := collected.actions ++ (a :: result.actions)
            state := result.state
            location := result.location
          }

def gpTightenAll : List Obj $\rightarrow$ Obj $\rightarrow$ Obj $\rightarrow$ State $\rightarrow$ GPBuild
  | [], l, _, s =>
      {
        actions := []
        state := s
        location := l
      }
  | n :: rest, l, m, s =>
      if s.statics.nut_t n = true $\wedge$
         s.dynamic.loose_p n = true then
        match gpFindCarriedUseableSpanner s m with
        | none =>
            gpTightenAll rest l m s
        | some sp =>
            let a := PlanAction.tighten_nut l sp m n
            let s' := actionApply a s
            let result := gpTightenAll rest l m s'
            {
              actions := a :: result.actions
              state := result.state
              location := l
            }
      else
        gpTightenAll rest l m s

def solve (s : State) (g : Goal) : List PlanAction :=
  let m := gpFindMan s
  let start := gpFindManLocation s m
  let traversed :=
    gpTraverse s.statics.objects.length start m s
  let tightened :=
    gpTightenAll s.statics.objects traversed.location m traversed.state
  traversed.actions ++ tightened.actions
\end{lstlisting}


\end{document}